\documentclass{article}

 \usepackage[preprint]{neurips_2026}

\usepackage[utf8]{inputenc} 
\usepackage[T1]{fontenc}    
\usepackage{hyperref}       
\usepackage{url}            
\usepackage{booktabs}       
\usepackage{amsfonts}       
\usepackage{nicefrac}       
\usepackage{microtype}      
\usepackage{xcolor}         

\usepackage{subcaption}

\usepackage{graphicx}
\usepackage{amsmath}
\usepackage{amssymb}
\usepackage{mathtools}
\usepackage{amsthm}
\usepackage[textsize=tiny]{todonotes}
\usepackage{subcaption} 
\usepackage{multirow}
\usepackage{arydshln} 
\usepackage{tabularx}
\usepackage{array}
\usepackage{makecell}
\usepackage{threeparttable}
\usepackage{longtable}
\usepackage{colortbl} 
\usepackage[acronym]{glossaries}
\usepackage{caption} 
\usepackage{wrapfig}
\usepackage{adjustbox} 
\usepackage{pifont} 
\usepackage{xcolor}
 
\usepackage{amsmath}
\usepackage{amssymb}
\usepackage{mathtools}
\usepackage{amsthm}
\usepackage{wrapfig}
\usepackage[shortlabels]{enumitem}
\setlist[enumerate]{nosep}

\makeglossaries

\newacronym{MAS}{MAS}{Multi-Agent Systems}
\newacronym{LLM}{LLM}{Large Language Models}
\newacronym{AI}{AI}{Artificial Intelligence}
\newacronym{RAG}{RAG}{Retrieval-Augmented Generation}
\theoremstyle{plain}
\newtheorem{theorem}{Theorem}[section]
\newtheorem{proposition}[theorem]{Proposition}

\theoremstyle{definition}
\newtheorem{definition}[theorem]{Definition}

\theoremstyle{remark}

\newtheorem{example}{Example}

\title{Responsible Agentic AI Requires Explicit Provenance}

\author{%
  Jinwei Hu$^{1}$, Xinmiao Huang$^{1}$, Qisong He$^{1}$, Youcheng Sun$^{2}$, Yi Dong$^{1}$, Xiaowei Huang$^{1}$ \\
  $^{1}$School of Computer Science and Informatics, University of Liverpool, UK \\
  $^{2}$Department of Computer Science, Mohamed bin Zayed University of Artificial Intelligence, UAE
}

\begin{document}

\maketitle

\begin{abstract}
Agentic AI is rapidly proliferating across diverse real-world domains such as software engineering, yet public trust has not kept pace. The central reason is that responsibility, despite being widely discussed, remains a subjective and unenforced concept, as no current agentic framework produces the quantifiable, traceable, and interventionable provenance needed to assign it when harm emerges from compositions no single party designed. We position that what is missing is not better benchmark-level evaluation but \textbf{explicit provenance} across the full agentic lifecycle, which is the only viable basis for making responsibility computable and actionable. We advance this agenda along four axes: establishing \textit{why} such provenance is a structural necessity by identifying responsibility gaps across sociotechnical dimensions, formalizing \textit{what} it must encode through a causal attribution function and responsibility tensor, discussing \textit{how} it can be made computable across four lifecycle layers with preliminary experiments showing that provenance is estimable and interveneable online before irreversible harm accumulates, and examining \textit{who} bears responsibility through a concrete agentic incident. Explicit provenance is not a discretionary refinement but the necessary condition for responsible agentic AI, and no stakeholder across its ecosystem can afford to treat it as optional.

\end{abstract}


\section{Introduction}
\label{sec:introduction}

\begin{figure}[t]
    \centering
    \vspace{-13pt}
    \includegraphics[width=\linewidth]{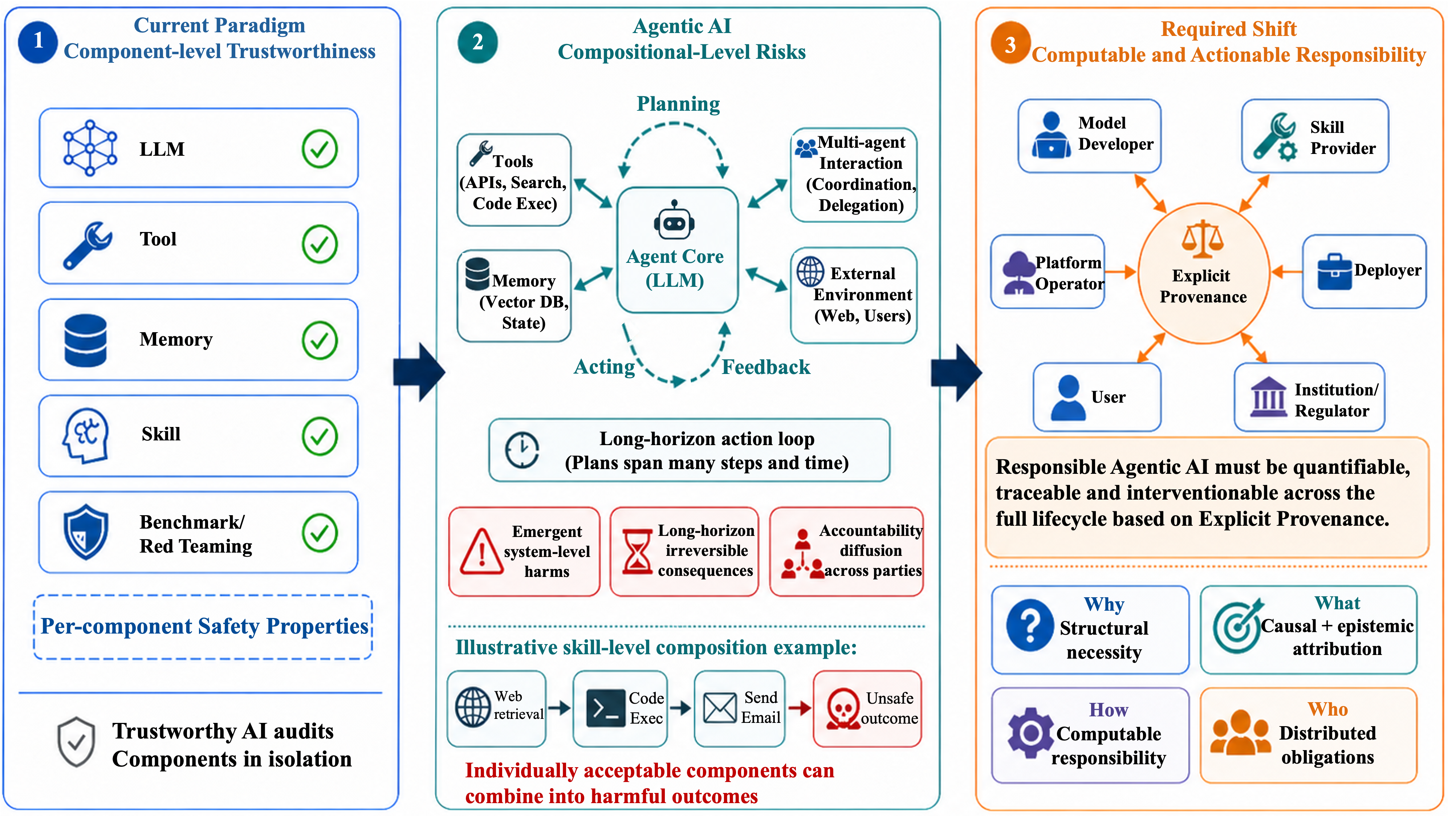}
    \caption{Per-component trustworthy AI audits components in isolation (\textbf{1}), but agentic AI embeds them in long-horizon action loops that produce emergent harms, unanticipated compositional consequences, and diffused responsibility (\textbf{2}). Closing this gap requires explicit provenance  across the full lifecycle of responsible agentic AI and we demonstrate it via four axes (\textbf{3}).}
    \label{fig:results}
    \vspace{-10pt}
\end{figure}

Artificial intelligence (AI) has recently undergone three transitions in the scope of its action. Early AI systems were largely reactive: inputs in, outputs out. A second generation, grounded in large language models (LLMs)~\citep{brown2020language,achiam2023gpt,touvron2023llama}, shifted the focus from prediction to open-ended text generation. A third transition is now underway: \emph{agentic AI}, in which models are embedded in autonomous action loops, equipped with tools and persistent memory, integrated into multi-agent pipelines, and granted authority to plan, execute, and revise decisions over extended horizons~\citep{wang2024survey,xi2025rise,jiang2026sok}. This transition changes the object of concern: an agentic system does not merely answer questions, but acts in the real world by booking flights, executing code, sending emails, and coordinating with other autonomous systems, often with little or no human intervention~\citep{park2023generative,hong2023metagpt,wu2024autogen,Hu_Dong_Sun_Huang_2026,li2026skillsbench}.

Yet despite this rapid proliferation, enterprise surveys report that real-world adoption remains far below the pace of development and organizations scale capability faster than they are willing to delegate authority~\citep{mckinsey2026trust,deloitte2026agentic}. The practical consequence is an adoption bottleneck rooted in a trust deficit. Users, organizations, and regulators hesitate to grant consequential authority to systems whose actions leave no traceable record of how decisions were reached, by whom, and through which causal chain. This is not irrational caution but a predictable response to structural unaccountability: when harm occurs, each party points to another and responsibility dissolves before it can be assigned. Responsibility has been widely discussed in the trustworthy AI literature~\citep{liu2022trustworthy,li2023trustworthy}, yet it has never been made computable or actionable across the full agentic lifecycle, not for lack of normative vocabulary but for lack of the explicit provenance. Consider a case where individually audited skills each pass safety review, yet their composition within an LLM-based agent produces regulatory violations that no per-skill audit could have detected: no party can be held accountable precisely because no responsibility envelope was specified before deployment to bound each party's obligations, and no record exists of how the harmful trajectory was jointly produced.

This failure is not incidental but structural, as harm in agentic systems does not reside in any single component but emerges from system-level trajectories, compositions, and distributed causal chains that no individual audit can inspect~\citep{chan2023harms}. Existing records are designed around per-component failures and therefore remain too coarse to support causal responsibility attribution or its traceability across parties, and their post-hoc nature means that intervention cannot be mounted until after harm has already occurred~\citep{ojewale2026audit}. \textbf{\textit{We therefore position that responsible agentic AI must be quantifiable, traceable and interventionable across the full lifecycle, and that explicit provenance is the necessary infrastructure for making the responsibility computable and actionable.}} Without it, responsibility remains an subjective concept that cannot be assigned, or enforced, and benchmark-driven evaluation, while necessary for component-level capability, neither reaches compositional behavior nor produces the causal attribution that the responsibility requires.

This paper therefore advances a provenance-grounded agenda for computable and actionable responsibility in agentic AI along four axes. We first establish \textbf{why} explicit provenance is a structural necessity rather than a discretionary refinement, identifying socio-technical responsibility gaps (Section~\ref{sec:why}). We then formalize \textbf{what} provenance must encode in agentic contexts through a causal attribution function and an responsibility tensor (Section~\ref{sec:what}). We address \textbf{how} responsibility can be made computable across four lifecycle layers, validated through preliminary experiments showing that causal contribution is estimable online before irreversible harm accumulates (Section~\ref{sec:how}). Finally, we examine \textbf{who} bears responsibility by applying the responsibility tensor to a concrete agentic incident, demonstrating how provenance distributes obligations (Section~\ref{sec:who}).

\section{Background}
\label{sec:background}

Agentic AI now appears across diverse domains, from software automation to professional decision support \citep{yao2022react, schick2024toolformer, zhou2024webarena, HU2025103779, xu2026theagentcompany}. Thus, we adopt a general functional definition centered on the capabilities that make these systems consequential.

\begin{definition}[Agentic AI System]
\label{def:agentic}
An agentic AI system situates one or more LLM-based agents within a perception-action loop with (i) \emph{tool access}, (ii) \emph{persistent memory}, (iii) \emph{multi-step planning}, and (iv) \emph{self-correction}, optionally coordinating with other agents.
\end{definition}

\textbf{Agentic adoption is outpacing responsibility structures.} Frontier providers are productizing agents that use tools, operate computers, and complete multi-step tasks, while open ecosystems such as OpenClaw and ClawHub illustrate the emergence of third-party skill marketplaces where agents can be extended with externally authored capabilities~\citep{openai2025operator,openai2025agents,anthropic2026tooluse,openclaw2026,clawhub2026}. Enterprise surveys confirm that adoption is already at scale: McKinsey reports that 23\% of respondents are scaling agentic AI systems and another 39\% experimenting, and PwC reports that 79\% of senior executives say AI agents are already being adopted in their companies~\citep{mckinsey2025stateai,pwc2025agents}. Yet responsibility structures remain less mature, with McKinsey identifying persistent gaps in governance and risk management and Deloitte characterizing agentic AI as scaling faster than its guardrails~\citep{mckinsey2026trust,deloitte2026agentic}. Agentic AI is therefore not merely a capability frontier but a responsibility frontier.
 
\textbf{Relation to trustworthy AI and adjacent work.} The trustworthy AI paradigm has organized safety around model-level properties, benchmark evaluation, and red-teaming~\citep{liangholistic,ganguli2022red,sun2023accountability}, but agentic deployment exposes three structural limitations. Risks is \emph{compositional}: properties verified for individual components need not survive their interaction~\citep{li2026skillsbench}. Evaluation is \emph{fragile}: task completion can diverge from safety, policy compliance, and harmfulness control~\citep{andriushchenkoagentharm,kuntzharm,hu2026lyingtruthsopenchannelmultiagent}. Accountability is \emph{diffuse}: harmful outcomes arise from multi-step trajectories distributed across developers, platforms, and deployers that no per-component audit can reconstruct~\citep{staufer20262025}. Existing accountability work provides foundations for auditing and institutional oversight~\citep{doshi2017accountability,wieringa2020account}, but assumes bounded decisions and inspectable systems that agentic AI renders obsolete: harm may emerge from long-horizon trajectories, persistent memory, and third-party skills that no single actor designed or controlled end-to-end. Responsibility discourse in moral philosophy, legal theory, and AI governance is extensive~\citep{vincent2011moral,porter2025unravelling,novelli2024accountability}, yet none of these bodies of work produces the explicit provenance that makes responsibility computable when harm emerges from compositional agentic behavior rather than from any identifiable single decision.

\section{Why Responsible Agentic AI Requires Explicit Provenance}
\label{sec:why}

Across the limitations identified in Section~\ref{sec:background}, responsibility remains uncomputable due to no party produces the explicit provenance needed to assign responsibility when harm emerges from individually compliant components. We therefore characterize what any responsible agentic system must satisfy.

\begin{proposition}[Structural Requirement of Responsible Agentic AI]
\label{prop:necessity}
A responsible agentic system requires explicit provenance satisfying three properties. First, \emph{quantifiability}, meaning it produces measurable causal attribution for each component and corresponding party. Second, \emph{traceability}, meaning it grounds responsibility assignments in causally verified execution records and epistemic position. Third, \emph{interventionability}, meaning it is produced continuously so that agentic systems remain open to interception and recovery across the lifecycle.
\end{proposition}

The following analysis demonstrates, for each property, why its absence leaves the corresponding sociotechnical dimensions unresolvable. These eight dimensions are representative rather than exhaustive, synthesized from the adjacent literature on responsible AI~\citep{vincent2011moral,matthias2004responsibility,vladeck2014machines,cobbe2023understanding,dafoe2018ai,hu2025stopreducingresponsibilityllmpowered,porter2025unravelling}.
 
\subsection{Sociotechnical Demonstration of Responsibility Gaps}
\label{sec:why:analysis}
 
\paragraph{Quantifiability.}
\textbf{Law} and \textbf{Morality} both require an identifiable causal agent whose contribution to a harm can be established. No party manufactures an agentic composition, so the EU AI Liability Directive's presumption of fault cannot attach when harm is a property of component interaction rather than any individual component~\citep{vladeck2014machines,euliability2022,balke2008conclusion,oliver2021contracting,cobbe2023understanding}. Morally, agentic causation is distributed across developers, skill maintainers, and platform operators in proportions no party designed or anticipated, so no party clears the culpability threshold and victims have no one to hold accountable~\citep{matthias2004responsibility,vallor2024find,cho2015survey}. At the technical level, \textbf{Standards} can make responsibility actionable only if they specify what causal attribution a system must produce to demonstrate responsible deployment; without this, every party defines responsible deployment on its own terms and no objective criterion exists to determine whether any obligation was met or breached~\citep{hu2025stopreducingresponsibilityllmpowered,raji2020closing}. \textbf{Regulation} faces the same dependency. Mandatory requirements can close the accountability gap only if they specify what quantifiable attribution parties must produce as a precondition for market participation, since in open agentic marketplaces compositional safety testing generates collective benefits but imposes private costs, making zero investment the individually rational outcome absent such requirements~\citep{dafoe2018ai,mckinsey2026trust,deloitte2026agentic}. Without quantifiability, no legal, moral, standards-based, or regulatory framework can identify which party contributed what influence, leaving all four dimensions permanently unresolvable regardless of how well individual components are audited.
 
\paragraph{Traceability.}
\textbf{Ethics} and \textbf{Values} frameworks are designed to prevent harms that accumulate gradually across interactions rather than materializing at any single step. Agentic systems accumulate effects across weeks or months that no per-interaction audit detects~\citep{gabriel2024ethics}: autonomy erodes as delegation narrows users' effective choices~\citep{mok2025exploring}, and compositional bias amplification compounds small discriminatory signals into population-scale outcomes invisible at any single step~\citep{ferrara2024genai,yin2023long,bommasani2022picking,lee2024large}. Without traceability that grounds cumulative causal attribution across interactions and populations, these harms remain structurally invisible to any responsibility framework. At the technical level, \textbf{Practice} failures are invisible to per-component auditing for the same reason: no traceable causal attribution of the joint trajectory is available. Indirect prompt injection succeeds with attack rates reaching 47\% against GPT-4 across 17 tool-integrated systems~\citep{zhan-etal-2024-injecagent}; over 68\% of sandbox scenarios exhibit potential real-world agent failures~\citep{ruanidentifying}; individually valid skills introduce composition risks that reduce task performance~\citep{li2026skillsbench}; and over 90\% of high-popularity skills in ClawHub fail security review~\citep{guo2026skillprobe}. In every case, the failure is a system-level property that only trajectory-level traceability can reconstruct and attribute.
 
\paragraph{Interventionability.}
\textbf{Professionalism} concerns the boundary beyond which delegating consequential decisions to an agentic system is not legitimate. In regulated domains such as medicine~\citep{habli2020artificial}, professional obligations cannot be transferred to an automated system on benchmark accuracy alone: benchmarks cover in-distribution tasks while deployment exposes systems to rare events and novel authority-boundary situations that benchmarks do not represent~\citep{yuan-etal-2024-r,lu2025out}. Without interventionability that continuously monitors whether automated authority remains within validated boundaries and intercepts execution when it does not, no mechanism can prevent irreversible harm from accumulating in the uncovered region. The same requirement applies to the gradual harms identified under \textbf{Ethics} and \textbf{Values}: harms that build across interactions without crossing any single detectable threshold can only be interrupted by a system that produces provenance continuously and acts on it in real time. Without interventionability, explicit provenance reduces to a forensic instrument and responsible operation across the lifecycle cannot be guaranteed.
 
\paragraph{Explicit Provenance as Necessary Infrastructure.} The socio-technical analysis above shows the three properties are jointly necessary rather than independently sufficient. Quantifiability without traceability leaves contested assignments unresolvable, traceability without interventionability confines attribution to post-hoc reconstruction, and interventionability without quantifiability produces continuous monitoring with no basis for assigning what was detected to any responsible party. Since real deployments routinely instantiate gaps across all three classes simultaneously, Proposition~\ref{prop:necessity} cannot be satisfied without explicit provenance as the underlying infrastructure that makes all three properties simultaneously operational.

\section{What Explicit Provenance Must Encode}
\label{sec:what}
This section formalizes what explicit provenance must capture to make sociotechnical responsibility computable in agentic AI. The central question is not only which parties bear responsibility when harm emerges from an agentic trajectory, but what causal and epistemic evidence must be produced and maintained to make that assignment possible. The formalization is guided by three requirements established in Proposition~\ref{prop:necessity} and the definitions below instantiate these requirements and provide the formal basis for computable responsibility assignment.

\subsection{Causal and Epistemic Grounding of Explicit Provenance}

Let $\mathcal{A}=\{a_1,\ldots,a_n\}$ be the set of agents, $\mathcal{S}=\{s_1,\ldots,s_m\}$ the set of skills or tools and $\mathcal{E}$ the environment state space. An \emph{agentic trajectory} is a finite sequence $\tau=(e_0,\alpha_0,e_1,\alpha_1,\ldots,e_T)$, where $e_t\in\mathcal{E}$ is the environment state and $\alpha_t=(\alpha_t^1,\ldots,\alpha_t^n)$ is the joint agent action at step $t$. Actions may include model outputs, tool calls, skill invocations, inter-agent messages, or external operations. A \emph{harm event} $\omega\in\Omega$ is a measurable consequence of $\tau$ that negatively affects stakeholders. Let $\mathcal{P} = \{p_i \to M_i\}_{i=1}^{r}$ denote the deployment-chain accountability dictionary over $r$ parties, where $p_i$ is a human party and $M_i$ is the set of technical components (e.g., foundation models, skills and tools) developed, deployed, or maintained by $p_i$. Since harm in agentic systems can emerge from an extended trajectory rather than a single decision, responsibility assignment first requires a measure of each deployment-chain party's causal contribution to the harmful outcome. Hence, producing and maintaining this measure is the first class of evidence that explicit provenance must encode.

\begin{definition}[Causal Contribution]
\label{def:causal}
Let $\Pr[\omega\mid\tau]$ denote the probability that harm event $\omega$ occurs under trajectory $\tau$. The causal contribution of party $p$ is
\[
\kappa(p,\omega,\tau)=\Pr[\omega\mid\tau]-\Pr[\omega\mid\tau_{-p}],
\]
where $\tau_{-p}$ denotes a counterfactual trajectory under a role-preserving intervention that removes or neutralizes $p$'s relevant decisions while holding other conditions fixed where meaningful. $\kappa(p,\omega,\tau)>0$ iff $p$ has nonzero causal contribution.
\end{definition}

\noindent\textit{Remark 4.1.}
$\kappa$ is a theoretical anchor rather than a directly observed quantity. Recent work on auditable and traceable LLM agents suggests that causal evidence can be approximated from execution traces, tool-activation records, permission boundaries, counterfactual replay, and controlled ablations~\citep{ojewale2026audit,pmlr-v267-zhang25cq,andriushchenkoagentharm,zhang2026agentracer}. The role of $\kappa$ is therefore not to assume perfect attribution, but to specify the kind of evidence that responsibility analysis should seek when attribution is contested. We provide preliminary evidence in Section~\ref{sec:nesy_trial_responsible_agentic_ai} that causal signal is estimable online from execution prefixes across heterogeneous agent environments, establishing the tractability of $\kappa$ at runtime. Attribution to human parties follows directly from the mapping in $\mathcal{P}$, in which each component is assigned to the party that developed, deployed, or maintained it.

Moreover, causal contribution alone is insufficient for responsibility assignment, as a party's responsibility also depends on what it knew, or should reasonably have known, before the harm occurred. Each party's epistemic position therefore constitutes the second class of evidence that explicit provenance must encode, and we define it using an objective standard so that parties cannot evade responsibility by manufacturing ignorance through incomplete documentation or inadequate safety assessment.

\begin{definition}[Epistemic Position]
\label{def:epistemic}
$\varepsilon_p^t=(\mathcal{I}_p^t,\mathcal{C}_p^t)$, where $\mathcal{I}_p^t$ is the information available or reasonably expected to be available to party $p$ at time $t$, and $\mathcal{C}_p^t\subseteq\Omega$ is the set of harm events $p$ could have reasonably foreseen given $\mathcal{I}_p^t$, assessed against the \emph{objective standard} of what a reasonably informed actor in $p$'s role should have anticipated. Party $p$ is \emph{epistemically culpable} for $\omega$ if $\omega\in\mathcal{C}_p^t$ for some $t$ prior to the harm.
\end{definition}

\subsection{From Explicit Provenance to Computable Responsibility}

The preceding definitions establish the causal and epistemic evidence that explicit provenance must encode, separating two conditions for responsibility into a universal function over responsible parties.

\begin{definition}[Computable Responsibility in Agentic System]
\label{def:responsibility}
The responsibility assignment function $\rho:\mathcal{P}\times\Omega\to[0,1]$ assigns an individual responsibility weight to each deployment-chain party $p$ for each harm event $\omega$. It satisfies:
\begin{enumerate}[leftmargin=1.6em,noitemsep,topsep=2pt]
\item \textbf{Proportionality.} 
$\rho(p,\omega)\propto\kappa(p,\omega,\tau)\cdot\mathbf{1}[\omega\in\mathcal{C}_p^t]$, where $\mathbf{1}[\omega\in\mathcal{C}_p^t]$ is an indicator function that equals $1$ if $\omega$ belongs to the set of harms reasonably foreseeable to party $p$ at time $t$.
\item \textbf{Completeness.} For every harm event $\omega$, $\sum_{p\in\mathcal{P}}\rho(p,\omega)+\rho_{\mathrm{inst}}(\omega)=1$, where $\rho_{\mathrm{inst}}(\omega)$ denotes residual institutional responsibility. When individual assignment is incomplete, the residual should be assigned to the institutional layer (Definition~\ref{def:institutional}).
\item \textbf{Non-evasion.} $\rho(p,\omega)>0$ whenever $\kappa(p,\omega,\tau)>0$ and $\omega\in\mathcal{C}_p^t$ for some $t$ prior to the harm. Thus, real and foreseeable contribution cannot be erased by system complexity.
\end{enumerate}
\end{definition}

\noindent\textit{Remark 4.3.}
Proportionality ties responsibility to causal contribution under reasonable foreseeability. Completeness assigns residual responsibility institutionally rather than leaving it unaccounted and serves as the normalization constraint that makes the proportionality relation precise. Non-evasion prevents foreseeable contributors from escaping responsibility through system complexity~\citep{matthias2004responsibility}. Together, these conditions make responsibility an allocative structure.

Finally, the institutional layer captures residual responsibility that cannot be exhausted by individual causal contribution and foreseeability. We define residual institutional responsibility as below:

\begin{definition}[Residual Institutional Responsibility]
\label{def:institutional}
Residual institutional responsibility is the portion of responsibility not assigned to individual deployment-chain parties:
$\rho_{\mathrm{inst}}(\omega)=1-\sum_{p\in\mathcal{P}}\rho(p,\omega)\geq 0$.
It is borne by the institutional layer when that layer had the authority and capacity to close or absorb the accountability gap through standards, certification, monitoring, incident response, or compensation mechanisms, but failed to do so.
\end{definition}

\subsection{Distributing Responsibility Across Deployment-Chain Parties}
Responsibility assignment requires decomposing an observed harm event $\omega$ into component-level causal contributions and mapping them back to the parties that bear them. We use the scalar $\rho(p,\omega)$ to quantify each component's individual contribution, and the deployment-chain dictionary $\mathcal{P}=\{p_i\to M_i\}_{i=1}^{r}$ to map each contributing component to its responsible party. Since agentic harm typically implicates multiple parties across multiple socio-technical dimensions simultaneously, we define the responsibility tensor to make this full allocation structure explicit and computable.
 
\begin{definition}[Responsibility Tensor]
\label{def:tensor}
Let $\mathcal{D}=\{d_1,\ldots,d_k\}$ be the set of socio-technical responsibility dimensions specified for the deployment context. The responsibility tensor $\mathbf{R}\in[0,1]^{|\mathcal{P}|\times|\Omega|\times|\mathcal{D}|}$ encodes dimension-specific responsibility, where $\mathbf{R}[p,\omega,d_k]$ denotes party $p$'s responsibility for harm event $\omega$ along dimension $d_k$. The scalar assignment is recovered as $\rho(p,\omega)=\sum_{k=1}^{|\mathcal{D}|} w_k\mathbf{R}[p,\omega,d_k]$, where $w_k>0$ reflects the relative significance of dimension $d_k$ in the deployment context, with $\sum_{k=1}^{|\mathcal{D}|}w_k=1$.
\end{definition}
 
The weights $w_k$ reflect the relative significance of each responsibility dimension and should be explicitly recorded in the agentic context. A key advantage of this formalization is that the tensor remains computable from $\mathcal{P}$ even without full provenance by reducing to party identification alone, ensuring responsible parties can always be identified; with explicit provenance, each entry is causally grounded, making responsibility attribution actionable across the lifecycle.


\section{How Explicit Provenance Makes Responsibility Computable}
\label{sec:how}

This section addresses how explicit provenance can be produced across the full agentic lifecycle. The central obstacle is that $\kappa(p,\omega,\tau)$ requires estimating $\Pr[\omega \mid \tau]$, a quantity not directly recoverable from raw execution logs without structured, failure-aligned provenance records. Making explicit provenance operational therefore requires progress across four sequentially dependent layers: L1 establishes the causal structure that L2 must instrument, L2 produces the execution records that L3 translates into normative judgments, and L3 establishes the accountability structure that L4 must extend to population scale. Without progress on all four layers, responsibility will remain a theoretical construct rather than an operational instrument.

\subsection{L1: Design}
\label{sec:how:design}

Quantifiability requires the causal structure of an agentic system be known before harm occurs~\citep{peters2017elements,geiger2021causal}. The explicit provenance needed to approximate causal signal from execution must be designed into the system, not merely relying on reconstructed post-hoc logs. We represent this structure as a multilevel dependency graph $\mathcal{G} = (\mathcal{V}, \mathcal{E})$, where $\mathcal{V}$ spans parties, models, skills, and their underlying components, and $\mathcal{E}$ encodes causal influence across levels. Current practice does not meet this requirement: skills, tools, agents, and parties are composed without documenting causal dependencies~\citep{li2026skillsbench,guo2026skillprobe}. The result is that $\Pr[\omega \mid \tau]$ is unestimable not because the mathematics fails but because the system was never designed to produce the provenance that quantifiability requires. The research agenda this motivates is \textbf{causal decomposition of compositional agentic systems}, representing multilevel dependency structure in a form that supports provenance production at any granularity, and establishing what constitutes a meaningful component boundary through interdisciplinary engagement with law, ethics, and domain expertise~\citep{dong2025safeguarding}.

\subsection{L2: Engineering}
\label{sec:how:engineering}

Even with a well-defined causal structure at design time, explicit provenance remains unproduced if the execution record does not capture the evidence required to evaluate $\kappa$ and $\varepsilon_p^t$. This engineering barrier has two dimensions corresponding to traceability and interventionability respectively. Traceability requires \textit{retrospective coverage}: without structured, failure-aligned representations of each execution step, the causal signal required to estimate $\Pr[\omega \mid \tau]$ cannot be recovered from raw logs, and the epistemic records needed to establish $\varepsilon_p^t$ are lost~\citep{ojewale2026audit}. Interventionability requires \textit{prospective coverage}: agentic systems routinely take irreversible actions, meaning that provenance available only after harm has materialized cannot support interception or recovery. Responsible agentic AI must therefore be equipped with the capacity to estimate trajectory risk in real time and interrupt execution before irreversible consequences accumulate. Both gaps map directly into unresolvable $\rho_{\mathrm{inst}}$.

The research agenda this layer motivates has two components that together address traceability and interventionability. First, \textbf{online trajectory monitoring and risk estimation}: agentic systems require monitors that produce structured, inspectable causal representations from execution traces in real time and estimate $\Pr[\omega \mid e_t, \alpha_{t:T}]$ before irreversible consequences accumulate, where $e_t \in \mathcal{E}$ is the current environment state and $\alpha_{t:T} = (\alpha_t, \ldots, \alpha_T)$ is the planned action sequence to deployment-specified horizon $T$. Such monitors must be both flexible enough to handle heterogeneous traces and inspectable enough to serve as evidence in attribution~\citep{rudin2019stop}. Neuro-symbolic approaches are a promising direction precisely because they combine representational flexibility with auditable intermediate reasoning states~\citep{garcez2023neurosymbolic,pmlr-v267-zhang25cq,zhang2026agentracer}. Second, \textbf{compositional verification of responsibility attribution}: whenever a new component $x_{\text{new}}$ enters the system, the shift in causal attribution $\Delta\kappa(p, \omega, \tau, x_{\text{new}}) = \kappa\!\left(p, \omega, \tau \mid_{\mathcal{G} \oplus x_{\text{new}}}\right) - \kappa(p, \omega, \tau \mid_{\mathcal{G}})$, where $\mathcal{G} \oplus x_{\text{new}}$ must be tested before co-activation is permitted, ensuring that quantifiability is maintained as the system grows~\citep{li2026skillsbench}.

\subsection{L3: Deployment}
\label{sec:how:deployment}

L1 and L2 establish the technical infrastructure for explicit provenance, making $\kappa(p,\omega,\tau)$ computable from execution records and $\varepsilon_p^t$ reconstructable from structured logs, but translating provenance into a fully populated responsibility tensor $\mathbf{R}[p,\omega,d_k]$ requires human judgment that technical systems cannot produce on their own. Each normative dimension of $\mathbf{R}$ depends on decisions about foreseeability, authority, and obligation that must be made and documented by human actors before deployment. The deployment layer is therefore where quantifiability and traceability are translated into actionable responsibility assignments, and where the institutional conditions for interventionability are established so that agentic systems have the authority and procedures to intercept and recover from harmful trajectories~\citep{novelli2024accountability}. How dimension weights $w_k$ should be specified and adapted as harm evidence accumulates remains an open research direction; precedents from safety-critical domains~\citep{wang2023freematch,angelopoulos2024conformal} suggest that data-driven weight updating is both feasible and necessary for contextually grounded responsibility allocation across diverse deployment contexts.

\begin{definition}[Deployment Readiness Condition]
\label{def:readiness}
A responsible agentic system satisfies the deployment readiness condition iff: (i) the causal dependency structure is documented at all component levels; (ii) compositional verification covers all component combinations with nonzero co-activation probability; (iii) all parties have established a responsibility envelope documenting their obligations and intervention boundaries to downstream parties, with dimension weights $w_k$ specified for the deployment context; (iv) incident response plans have been exercised with human participants; (v) informed consent mechanisms have been validated with users.
\end{definition}

\subsection{L4: Experience}
\label{sec:how:experience}

The preceding layers make $\kappa$ computable for harm events at identifiable trajectory points, but do not address harm that builds up gradually across interactions without any single event crossing a detectable threshold, the mechanism identified as autonomy erosion and compositional bias amplification~\citep{gabriel2024ethics}. The relevant counterfactual is not what would have happened in this trajectory without $p$, but how the user's beliefs and choices would have differed without the system's persistent influence, which requires tracking divergence across trajectories at population scale and cannot be answered by per-interaction audit alone~\citep{bommasani2022picking}. This gap extends all three provenance properties beyond what L1 through L3 reach, as quantifiability must cover cumulative causal contributions across interactions, traceability must ground assignments in population-level execution records, and interventionability must support recovery before gradual harms become irreversible at scale. This motivates two complementary research directions. The first is population-scale monitoring of how much the system narrows users' actual choice sets relative to a counterfactual baseline, aggregated in ways sensitive to distributional harm, with thresholds for triggering $\rho_{\mathrm{inst}}$ jointly specified by technical and social science communities~\citep{bommasani2022picking}. The second is populating the user-relevant dimension of $\mathbf{R}$ on evidential rather than assumed grounds, since $\mathcal{C}^t_{\mathrm{user}}$ is non-empty only when meaningful risk information was disclosed before interaction, and responsibility regimes should not require technical literacy beyond what a reasonable user can be expected to have~\citep{doshi2017accountability}. The first captures cumulative harm that falls outside individual $\kappa$ assignments, while the second ensures that user-side entries in $\mathbf{R}$ reflect what users were actually in a position to foresee.

\subsection{Neuro-Symbolic Trial for Explicit Provenance}
\label{sec:nesy_trial_responsible_agentic_ai}
\begin{wrapfigure}{r}{0.68\textwidth}
\vspace{-23pt}
    \centering
    \includegraphics[width=0.68\textwidth]{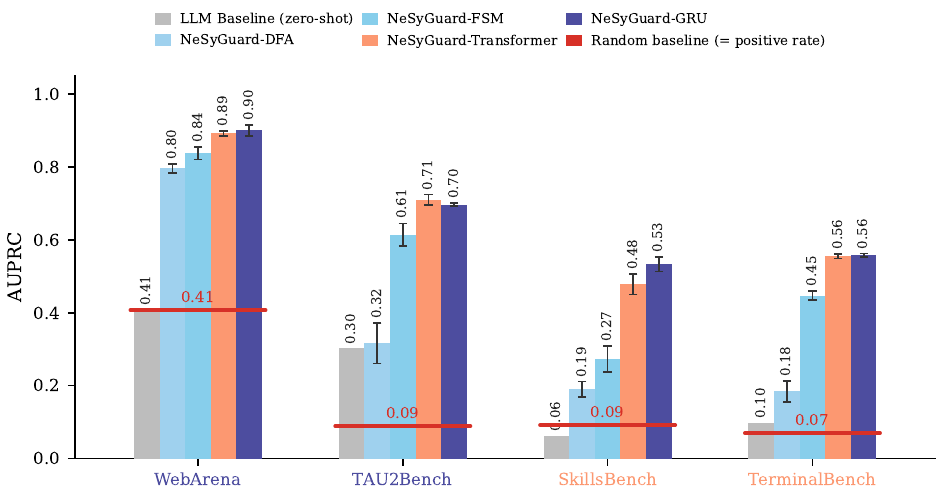}
    \vspace{-10pt}
    \caption{\textbf{Online causal signal is estimable from agent execution prefixes, supporting required properties for responsible agent.} AUPRC measures how well a monitor identifies failing trajectories before harm materializes. NeSy monitors substantially outperform random and zero-shot LLM baselines.}
    \label{fig:auprc_interpretation}
    \vspace{-15pt}
\end{wrapfigure}
We report preliminary experiments targeting the L2 with detailed implementation details in Appendix~\ref{app:implementation}, where online provenance signal is attributed from execution prefix traces to support responsible decision making across multi-component agentic systems. We study this through a plug-in neuro-symbolic (NeSy) monitoring architecture that attaches to agentic systems without modifying their internals. Raw agent traces are converted into a canonical step-level representation via a one-time offline adapter induction, compressed into learned failure-aligned event abstractions, and scored by monitors ranging from neural sequence models to auditable finite-state variants without deployment-time LLM judging. This design means that any agentic system producing execution logs can be equipped with provenance-based monitoring capacity, instantiating the computable responsibility.

Figure~\ref{fig:auprc_interpretation} shows AUPRC results across WebArena \citep{zhou2024webarena}, TAU2Bench \citep{barres2025tau2bench}, SkillsBench \citep{li2026skillsbench}, and TerminalBench \citep{merrill2026terminal}, where the red baseline reflects each benchmark's positive-prefix rate under random scoring. Three findings bear directly on the provenance agenda. All learned monitors exceed the random baseline by substantial margins, establishing that $\Pr[\omega \mid e_t, \alpha_{t:T}]$ is estimable online and that the quantifiability requirement is tractable at runtime. Finite-state variants compress warning behavior into symbolic states legible to human auditors, providing the inspectable causal evidence that traceability requires and that the epistemic position $\varepsilon^t_p$ in Definition~\ref{def:epistemic} depends on. Monitors further identify high-risk prefixes early enough to support intervention before irreversible consequences accumulate, satisfying the interventionability requirement that provenance be produced continuously.

\section{Who Bears Responsibility in Agentic System}
\label{sec:who}

\begin{wraptable}{r}{0.55\textwidth}
\vspace{-15pt}
\centering
\caption{WebArena DFA prefixes decomposed into interpretable states as explicit provenance. Each state reports failure-risk evidence (Risk), supporting-prefix count (Eval), normalized timing $\bar{t}/T$, and representative states for responsibility interpretation, support responsibility assignment or intervention.}
\label{tab:symbolic_illustration}
\vspace{-5pt}
\tiny
\setlength{\tabcolsep}{2pt}
\begin{tabular}{clrrcl}
\toprule
State & Behavioral Phase         & Risk  & Eval & $\bar{t}/T$ & Representative State                        \\
\midrule
\multicolumn{6}{l}{\textit{Warning states (risk $\geq$ 0.34)}}                                             \\
q0    & Early navigation reset   & 0.857 & 544  & 0.25        & \texttt{click; goto homepage}              \\
q28   & Explicit error message   & 0.548 & 40   & 0.81        & \texttt{type [out of stock\ldots]}         \\
q22   & Repetitive click loop    & 0.518 & 595  & 0.40        & \texttt{click$\times$6 (no type)}          \\
q12   & Misaligned search query  & 0.510 & 643  & 0.25        & \texttt{type [CMU]} \\
q24   & External-search redirect & 0.434 & 276  & 0.40        & \texttt{new\_tab; goto google.com;}   \\
q1    & Early scroll-and-click   & 0.342 & 379  & 0.25        & \texttt{click; scroll [down]}              \\
\midrule
\multicolumn{6}{l}{\textit{Representative normal states (risk $<$ 0.25)}}                                  \\
q17   & Productive backtracking  & 0.085 & 56   & 0.83        & \texttt{go\_back$\times$5}                 \\
q4    & Credential entry         & 0.099 & 139  & 0.67        & \texttt{type [username]; click}            \\
q26   & Task-specific search     & 0.038 & 90   & 0.50        & \texttt{type [color utility]; click}       \\
q8    & Long-form text entry     & 0.122 & 80   & 0.74        & \texttt{type [multi-sentence msg]}     \\
q7    & Short-label selection    & 0.119 & 111  & 0.75        & \texttt{type [feature]; click}             \\
\bottomrule
\end{tabular}
\vspace{-15pt}
\end{wraptable}
Section~\ref{sec:how} establishes that explicit provenance makes causal signal estimable online, but that answers only how much causal contribution a party made. Populating the full responsibility tensor $\mathbf{R}[p,\omega,d_k]$ requires mapping provenance signals to specific parties, answering not only how much but who bears responsibility. We illustrate how provenance-grounded symbolic states can structure this mapping through interpretable examples using a Deterministic Finite Automaton (DFA) monitor in Table \ref{tab:symbolic_illustration} and Appendix~\ref{tau2bench}.

\begin{example}[Responsibility Assignment from Provenance in WebArena]
\label{ex:webarena}
An agent tasked with multi-step web navigation is monitored online by the DFA from Section~\ref{sec:nesy_trial_responsible_agentic_ai}. As shown in Table~\ref{tab:symbolic_illustration}, the provenance states identify warning states $q_{12}$ (misaligned search query, risk $0.510$) and $q_{24}$ (external-search redirect, risk $0.434$) at $\bar{t}/T = 0.25$ and $0.40$ respectively, well before the task concludes. Both states surface early boundary violations and their representative trace evidence directly maps responsible components back to accountable parties through $\mathcal{P}$.
\end{example}

Taking these states as provenance evidence, $\mathbf{R}$ can be partially populated as follows. State $q_{12}$ points to the skill developer's failure to specify retrieval boundaries, contributing to $\mathbf{R}[\text{skill developer}, \omega, d_{\text{Standards}}]$ and $\mathbf{R}[\text{skill developer}, \omega, d_{\text{Practice}}]$. State $q_{24}$ points to the platform operator's failure to enforce navigation constraints, contributing to $\mathbf{R}[\text{platform operator}, \omega, d_{\text{Regulation}}]$ and $\mathbf{R}[\text{platform operator}, \omega, d_{\text{Professionalism}}]$. The agent developer bears weight on $d_{\text{Standards}}$ for deploying without compositional verification. The end user's entries remain negligible absent prior authorization or risk disclosure~\citep{weidinger2022taxonomy}, and residual responsibility falls to $\rho_{\mathrm{inst}}$ under Definition~\ref{def:institutional}. If dimension weights $w_k$ have been established as part of the responsibility envelope at L3, responsibility shares become further computable and precisely allocated across parties.

This example demonstrates that provenance-grounded states make responsibility assignment across deployment-chain parties tractable, transforming the question of who bears responsibility from an unanswerable subjective concept into a actionable and computable allocation.
\section{Alternative Views}
\label{sec:altviews}
 
\textbf{Provenance requirements impose overhead that constrains capability and slows deployment.} Requiring explicit provenance records across the full deployment chain adds instrumentation and verification costs that favor well-resourced actors and may slow deployment of beneficial systems~\citep{anderljung2023frontierairegulationmanaging}. Our position is that this objection misidentifies the bottleneck. The constraint on agentic AI deployment is not the absence of provenance infrastructure but the absence of trust~\citep{mckinsey2026trust,deloitte2026agentic}. Research on autonomous vehicles shows that public trust, not technical capability, is the binding constraint on adoption, as cities that prioritize safety and transparency accelerate deployment while those that do not face sustained public resistance regardless of how capable the underlying technology becomes~\citep{10.3389/fpsyg.2022.976023}. Provenance infrastructure does not cap what agentic systems can do; it builds the institutional trust that makes large-scale delegation of authority viable.
 
\textbf{Distributed responsibility produces irresolvable conflict between parties with incompatible interests and values.} Different actors in the deployment chain hold different commercial, cultural, and epistemic positions, and any responsibility assignment will be contested. The deeper concern is that no single weighting of normative dimensions in $\mathbf{R}$ can claim universal legitimacy across communities with genuinely different values. However, we posit that value pluralism is not a reason to abandon structured accountability but a reason to build accountability structures that are explicit about where value choices are made~\citep{novelli2024accountability,wieringa2020account}. The responsibility tensor $\mathbf{R}$ is designed precisely for this, as the dimension weights $w_k$ are not fixed by the framework but determined by domain-specific governance, making value disagreements legible and contestable rather than hidden inside opaque attribution decisions. Conflict over $w_k$ is the mechanism through which pluralistic norms are negotiated and actionable, not evidence that the framework cannot work~\citep{hu2025stopreducingresponsibilityllmpowered}.
\section{Conclusion}
\label{sec:conclusion}

Agentic AI is already acting in the world, yet the responsibility structures needed to govern it remain absent. This paper has positioned explicit provenance as the necessary infrastructure for making responsibility computable and actionable across the full agentic lifecycle. Through formal analysis, a four-layer agenda, and preliminary experiments, we have demonstrated that responsibility satisfying quantifiability, traceability, and interventionability is both necessary and tractable. The work ahead spans design, engineering, deployment, and population-scale experience, and requires sustained engagement with law, ethics, and the social sciences not as adjacent disciplines but as co-designers of the accountability infrastructure that agentic AI demands. Only by making responsibility computable and actionable can the field earn the public trust that large-scale delegation of authority requires. The technology is already moving faster than the responsibility structures that should govern it, and closing that gap before it becomes irreversible is a shared obligation across the agentic AI ecosystem.

\newpage
\bibliographystyle{plain}
\bibliography{ref}
\clearpage
\appendix
\section{Implementation Details of Neuro-Symbolic Trial}
\label{app:implementation}

This appendix describes the implementation of the neuro-symbolic monitoring trial reported in Section~\ref{sec:nesy_trial_responsible_agentic_ai}. The trial is a preliminary instantiation of the L2 engineering agenda designed to demonstrate the feasibility of the research directions for responsible agentic system proposed in this paper. It is not intended as a standalone system contribution; its purpose is to establish that online provenance signals satisfying quantifiability, traceability, and interventionability are producible from heterogeneous agent execution traces. 


\paragraph{Prefix and Warning Label.}
A trajectory is an ordered sequence of execution steps recorded during an agent's task attempt. A prefix is the partial observation available to the monitor up to a given step during online execution, representing what has been observed so far without access to future steps. Each trajectory is associated with a binary outcome indicating task success or failure as determined by a task-specific verifier. Given a fixed warning horizon, a prefix is labeled a positive warning target if and only if the trajectory fails and fewer than a fixed number of steps remain before termination; all other prefixes are labeled negative. This labeling captures the steps immediately preceding failure, where failure-relevant provenance signals are most concentrated.

\paragraph{Trace view.}
The monitor does not consume raw logs directly. Each benchmark trace is first mapped into a fixed StepView record with fields for metadata, observation, action, tool, arguments, result, and status. The adapter is induced once from training traces and then frozen, so validation and test traces are processed by deterministic code rather than by deployment-time LLM calls. Validation and test traces are not used during adapter induction; the LLM acts only as a design-time parser proposer, not as a runtime evaluator.

\paragraph{Event abstraction.}
Each StepView record is serialized in a fixed field order and encoded with a training-only TF-IDF vocabulary. A small differentiable projection maps the encoded step into a learned event alphabet. This alphabet is optimized for the prefix-warning objective, so the symbols are intended to capture recurring failure-relevant evidence rather than manually named task states.

\paragraph{Monitor variants.}
The same learned event stream is used by four online monitor variants that span different points on the neural--symbolic tradeoff. The Gated Recurrent Unit (GRU) and Transformer variants act as flexible neural prefix scorers. The soft Finite State Machine (FSM) variant keeps a distribution over latent finite states during neural deployment, exposing a structured intermediate representation. The Deterministic Finite Automaton (DFA) variant is extracted post-hoc from hard event assignments and serves as an audit diagnostic, testing whether the learned behavior can be compressed into compact, risk-separating symbolic states that are legible to human auditors and suitable as causal evidence in responsibility attribution.

\paragraph{Evaluation metric.}
For every partial trajectory prefix, the monitor emits a risk score for near-term failure under a fixed warning horizon $H$. Prefixes from failed trajectories inside the final $H$ steps are labeled positive; successful prefixes and earlier failed-trajectory prefixes are labeled negative. Area Under the Precision-Recall Curve (AUPRC) is used as the primary ranking metric because warning targets are sparse and positive-prefix rates vary substantially across benchmarks. AUPRC measures how well a monitor ranks high-risk prefixes above low-risk ones using continuous risk scores, with the random baseline equal to each benchmark's positive-prefix rate. A monitor that substantially exceeds this baseline demonstrates that causal signal is recoverable from execution prefixes online, directly supporting the requirement of Proposition~\ref{prop:necessity}.

\paragraph{Baselines.}
Three comparison conditions are included. The random baseline reflects the positive-prefix rate of each benchmark and represents the performance of a monitor with no predictive signal. The zero-shot Large Language Model (LLM) baseline applies a frontier language model as a full-prefix judge without any task-specific training, representing the upper bound of deployment-time LLM judging under matched evaluation conditions. The supervised neural controls use the same StepView representations and prefix labels as the learned monitors but without the discrete event abstraction layer, isolating the contribution of failure-aligned symbol learning. All conditions share the same prefix-labeling protocol and held-out evaluation splits to ensure comparability. The finite-state results should be read as an interpretability diagnostic rather than as the only deployable form of the method.

\section{Extended Provenance Example on Dialogue Agent Responsibility Attribution}
\label{tau2bench}

To demonstrate that provenance-grounded symbolic states are not specific to web navigation environments, Table~\ref{tab:dfa_state_alignment_tau2} reports DFA state behavioral provenance on TAU2Bench, a tool-using dialogue benchmark involving customer service agents. The extracted automaton yields 20 states; Table~\ref{tab:dfa_state_alignment_tau2} reports the three trusted warning states and six representative normal states, with the remaining states excluded due to insufficient calibration support. Each trusted state is characterized by its risk score, the number of prefixes routed to it, its normalized timing $\bar{t}/T$, and a representative execution step that anchors its behavioral interpretation.

Three trusted warning states surface coherent failure-precursor patterns. State $q_1$ (risk 0.357, $\bar{t}/T$ = 0.57) captures mid-dialogue grounded lookup fan-out, where repeated \texttt{get\_details} or reservation queries after the task is underway indicate difficulty resolving the customer request. State $q_{19}$ (risk 0.337, $\bar{t}/T$ = 0.59) captures out-of-policy handoff, where a special-case request such as post-booking insurance is refused or transferred as indicated by \texttt{respond [insurance not allowed]}. State $q_{15}$ (risk 0.218, $\bar{t}/T$ = 0.78) captures late unresolved troubleshooting, where refund-policy or service failures such as \texttt{respond [refund/MMS fails]} remain open near trajectory end. Normal states correspond to routine task phases such as initial greeting and identity lookup ($q_0$, risk 0.031), billing remediation ($q_{16}$, risk 0.082), and transactional updates ($q_{14}$, risk 0.060). Compared to WebArena, the TAU2Bench automaton is more concentrated, with $q_0$ alone routing 16,754 prefixes, reflecting the more structured nature of dialogue-based task execution. These results show that the neuro-symbolic monitoring architecture surfaces interpretable provenance states across heterogeneous agentic environments, and that the representative trace evidence within each state directly supports mapping responsible components back to accountable parties through $\mathcal{P}$, satisfying the traceability requirement that explicit provenance must provide inspectable causal evidence legible to human auditors.

\begin{table}
  \centering
  \caption{TAU2Bench DFA prefixes decomposed into interpretable states as explicit provenance. Each state reports failure-risk evidence (Risk), supporting-prefix count (Eval), normalized timing $\bar{t}/T$, and representative trace evidence for responsibility interpretation. Trusted warning states are listed first to support responsibility assignment or intervention.}
  \label{tab:dfa_state_alignment_tau2}
  \small
  \setlength{\tabcolsep}{3pt}
  \begin{tabular}{clrrcl}
    \toprule
    State & Phase                           & Risk  & Eval  & $\bar{t}/T$ & Representative Step                      \\
    \midrule
    \multicolumn{6}{l}{\textit{Trusted warning states}}                                                              \\
    q1    & Grounded lookup fan-out         & 0.357 & 3949  & 0.57        & \texttt{get\_details / reservation}      \\
    q19   & Out-of-policy handoff           & 0.337 & 31    & 0.59        & \texttt{respond [insurance not allowed]} \\
    q15   & Late unresolved troubleshooting & 0.218 & 456   & 0.78        & \texttt{respond [refund/MMS fails]}      \\
    \midrule
    \multicolumn{6}{l}{\textit{Representative trusted normal states}}                                                \\
    q0    & Greeting / identity lookup      & 0.031 & 16754 & 0.25        & \texttt{respond; customer lookup}        \\
    q11   & Policy summary / confirmation   & 0.137 & 2992  & 0.72        & \texttt{respond [summary]}               \\
    q12   & Info collection / guidance      & 0.116 & 1471  & 0.72        & \texttt{get\_order; respond}             \\
    q16   & Billing remediation             & 0.082 & 607   & 0.68        & \texttt{check/make payment}              \\
    q14   & Transactional update            & 0.060 & 411   & 0.81        & \texttt{send\_certificate; check\_sim}   \\
    q5    & Mid-course telecom diagnosis    & 0.018 & 42    & 0.38        & \texttt{toggle\_airplane\_mode}          \\
    \bottomrule
  \end{tabular}
\end{table}





\end{document}